\documentclass[journal]{IEEEtran}
\usepackage[dvips]{graphicx}
\graphicspath{{../pdf/}{./source/}}
\DeclareGraphicsExtensions{.pdf,.jpeg,.png}
\usepackage{multirow}
\usepackage{amsmath,amssymb,bm}
\usepackage{amsfonts,dsfont,color,bbm}
\usepackage{algorithm}
\usepackage{algorithmic}
\usepackage[font=footnotesize,labelfont=bf,justification=justified,singlelinecheck=false]{caption}
\usepackage{slashbox, subfigure}
\usepackage{cite,epstopdf,epsf}
\usepackage{url}
\usepackage{cleveref}

\usepackage[utf8]{inputenc}

\crefname{equation}{}{}

\usepackage{amsthm}

\newtheorem{remark}{Remark}

\usepackage{setspace}

\newcommand{\be}{\begin{equation}}
\newcommand{\ee}{\end{equation}}
\newcommand{\bea}{\begin{eqnarray}}
\newcommand{\eea}{\end{eqnarray}}

\newcommand{\Real}{{\mathbbm{R}}}
\DeclareMathOperator*{\argmax}{arg\,max}

\DeclareMathOperator*{\E}{\mathbb{E}}
\newcommand{\Lagr}{\mathcal{L}}
\DeclareMathOperator{\sign}{sign}

\AtBeginDocument{
\addtolength{\abovedisplayskip}{-0.5ex}
\addtolength{\abovedisplayshortskip}{-0.4ex}
\addtolength{\belowdisplayskip}{-0.5ex}
\addtolength{\belowdisplayshortskip}{-0.4ex}
\addtolength{\belowcaptionskip}{-3.6ex}
}

\begin{document}

\title{Spatio-temporal Sequence Prediction with Point Processes and Self-organizing Decision Trees}

\author{Oguzhan Karaahmetoglu and Suleyman S. Kozat \textit{Senior Member, IEEE}\thanks{This work is supported in part by Outstanding Researcher Programme Turkish Academy of Sciences.}
\thanks{Suleyman S. Kozat and O. Karaahmetoglu are with the Department of Electrical and Electronics Engineering, Bilkent University, Bilkent, Ankara 06800, Turkey, Tel: +90 (312) 290-2336, Fax: +90 (312) 290-1223, {(contact e-mail: \{kozat, koguzhan\}@ee.bilkent.edu.tr).}}
\thanks{Suleyman S. Kozat and O. Karaahmetoglu are also with the DataBoss A.S., Bilkent Cyberpark, Ankara 06800, (email: \{serdar.kozat, oguzhan.karaahmetoglu\}@data-boss.com.tr)} }

\maketitle
\begin{abstract}
	We study the spatio-temporal prediction problem and introduce a novel point-process-based prediction algorithm. Spatio-temporal prediction is extensively studied in Machine Learning literature due to its critical real-life applications such as crime, earthquake, and social event prediction. Despite these thorough studies, specific problems inherent to the application domain are not yet fully explored. Here, we address the non-stationary spatio-temporal prediction problem on both densely and sparsely distributed sequences. We introduce a probabilistic approach that partitions the spatial domain into subregions and models the event arrivals in each region with interacting point-processes. Our algorithm can jointly learn the spatial partitioning and the interaction between these regions through a gradient-based optimization procedure. Finally, we demonstrate the performance of our algorithm on both simulated data and two real-life datasets. We compare our approach with baseline and state-of-the-art deep learning-based approaches, where we achieve significant performance improvements. Moreover, we also show the effect of using different parameters on the overall performance through empirical results, and explain the procedure for choosing the parameters.

\end{abstract}
\begin{keywords}
Spatio-temporal Point Process, Earthquake Prediction, Crime Prediction, Hawkes Process, Adaptive Decision Trees, non-stationary Time-Series Data, Online Learning.
\end{keywords}

\section{Introduction}\label{sec:introduction}

\subsection {Preliminaries}

	Effective processing of spatio-temporal data carries vital importance due to a wide range of problem setups for applications such as sequence prediction, dynamic system modeling, and data assimilation \cite{8486964}, \cite{roberto1992seismic}, \cite{8937810}. These setups are frequently studied due to their critical applications like criminal and social event prediction, and predictive maintenance \cite{roberto1992seismic}, \cite{wang2019deep}, \cite{9078883}, \cite{zhu2019learning}. Spatio-temporal sequences consist of samples ordered by their occurrence time and tagged along with their exact locations in a 2-D space. The aim is to predict the number of events that will take place in future spatio-temporal intervals or estimate the precise location and time of the next event using the past event times and positions up to an observation point. Accurately forecasting the distribution of criminal events or precisely estimating the time and location of the next event could save many lives and expenses \cite{6781624}, \cite{8937810}, \cite{wang2019deep}. However, certain difficulties such as the non-stationary dynamics or sparse distribution of the events restrict the application of standard approaches to this problem directly \cite{du2016recurrent}, \cite{yang2018recurrent}. Thus, we present a point-process-based approach to address these issues. Our algorithm can partition the data into subregions, where non-stationary dynamics of event sequences are modeled in each subregion with individual but interacting processes.

    Deep learning-based forecasting approaches have been shown to be superior compared to other methods. Particularly, Recurrent Neural Network (RNN) achieved exceptional performances in many domains \cite{du2016recurrent}, \cite{yang2018recurrent}, \cite{wang2017spatiotemporal}. These significant improvements are due to the inherent memory structure in RNN models \cite{elman1990finding}. RNN also has two other variants, namely, Gated Recurrent Unit (GRU) and Long-Short Term Memory (LSTM) \cite{lstm}. These variants improve the state transition mechanism of the standard RNN model by introducing gate mechanisms that prevent exploding or vanishing gradients problems \cite{lstm}. Convolutional Neural Networks (CNN) are also applied to the same problem setup \cite{duan2017deep}, \cite{yu2017spatiotemporal}, \cite{gao2019eeg}. CNNs are specifically tailored to capture complex spatial patterns in data, which plays a crucial role in predicting the number of future events \cite{wang2017deep}, \cite{wang2017spatiotemporal}. Despite the significant performance improvements, deep architectures can only process structured data, i.e., fixed sampling intervals and discretized spatial locations. Therefore, additional pre-processing and post-processing stages are added to process data \cite{du2016recurrent}, \cite{mei2017neural}, \cite{wang2017deep}.
    
    Another line of studies focus on applying the point processes and statistical time-series models to the same setup \cite{du2016recurrent}, \cite{yang2018recurrent}, \cite{mei2017neural}. These approaches aim to model data directly, and they are specifically designed for the application domain \cite{cox1980point}, such as the Hawkes process with an intensity that is a function of past event times. Despite their success in modeling events, standard point process intensities are formulated only for the temporal domain \cite{cox1980point}, \cite{daley2007introduction}. Certain studies introduce spatio-temporal point processes by extending the formulation for the spatial domain \cite{du2016recurrent}. Thus, allowing the estimation of the location of an event along with its occurrence time. Moreover, certain approaches present deep learning-based formulations for the intensity function \cite{du2016recurrent}, \cite{yang2018recurrent}.
    
    Here, we introduce a novel spatio-temporal prediction algorithm. We model the data as a non-stationary sequence in both time and space, as is the case in many real-life applications \cite{mei2017neural}, \cite{kanamori2003earthquake}. We do not assume any prior information about the sparsity of the data, which makes our approach applicable to different domains. Our method is based on point-processes, where we extend the formulation for the spatio-temporal sequences. We adaptively partition the space into subregions and model sequences in each subregion with interacting processes. Although we focus on the intensity function of the Hawkes process, our formulation can be readily extended to different intensity formulations. We show that our algorithm can model event sequences in real-life data such as crime and earthquake datasets. Finally, we provide a gradient-based optimization procedure for likelihood-maximization. Through this optimization procedure, we jointly optimize the partition boundaries and interacting point processes.

\subsection {Prior Art and Comparisons}

	Deep neural networks have been applied to many real-life prediction problems due to their abilities to model complex patterns \cite{elman1990finding}, \cite{wang2017deep}, \cite{krizhevsky2012imagenet}. Particularly, RNNs demonstrate exceptional performance in modeling temporal patterns due to their inherent memory \cite{elman1990finding}. Although deep models can model nonlinear patterns, their performance depends on the availability of massive amounts of labeled data. Moreover, their time-invariant formulation limits their capability under non-stationary sequences \cite{du2016recurrent}, \cite{wang2016isotonic}. The samples in spatio-temporal data can be unevenly spaced in time and continuously distributed in space, whereas deep models process structured data \cite{wang2017spatiotemporal}, \cite{du2016recurrent}. Additional pre-processing and post-processing stages may be designed for such cases \cite{wang2017deep}, \cite{wang2018graph}. However, these stages change the problem description as the continuous nature of the event times and locations are changed. Furthermore, model estimations will also be generated for the fixed spatio-temporal bins \cite{wang2018graph}. Unlike this approach, we directly model the continuous data without applying any ad-hoc pre-processing stage.

    Point processes are also applied to the same problem \cite{mei2017neural}, \cite{wang2016isotonic}, \cite{mohler2011self}. However, standard point processes are developed for modeling event occurrence times only, which prevents their direct application on spatio-temporal sequences \cite{cseke2016sparse}, \cite{adepeju2016novel}. To mitigate this problem, marked point processes were developed. Marked point processes estimate additional tags such as the location or type of the event along with its time \cite{jacobsen2006point}. Although this method has shown outstanding performance in modeling real-life trajectory sequences, it independently estimates the temporal and spatial locations. There are also studies that combine deep models with point processes, as in \cite{mei2017neural}, \cite{du2016recurrent}. These methods can capture nonlinear temporal patterns due to the capabilities of deep recurrent models. However, they incorporate auxiliary information including the location of events to the intensity function additively, which limits the generalization potential of the formulation. 
    
    Here, we address these issues by formulating a point-process intensity that is a function of both time and space. Instead of incorporating location and time information additively, we use a temporal and a spatial kernel mechanism along with an interaction mechanism to model dependencies in adaptive spatial subregions. Finally, we give a gradient-based likelihood-maximization procedure that jointly optimizes all model parameters.

\subsection {Contributions}

Our main contributions are:

\begin{enumerate}
		
	\item As the first time in the literature, we present a novel algorithm that adaptively partitions the spatial region into subregions and model the interaction between these subregions jointly.
	
	\item Although our formulation focuses on the self-exciting Hawkes process, our approach is generic so that any other point process can be used depending on the application as provided remarks in the paper.
	
	\item We provide a gradient-based optimization procedure for parameter inference. We use a log-likelihood-based objective function, which can be optimized sequentially in both online and batch setups.
	
	\item Through an extensive set of experiments on both simulated and real-life data, we show that our model can represent a spatio-temporal data such as earthquake and crime data, which are highly non-stationary. We compare our approach with the standard well-known methods where we demonstrate significant performance improvements.

\end{enumerate}

\subsection {Organization of the Paper}

	In the following section, we introduce our problem description. In section III-A, we briefly describe the probabilistic models that we use. In section III-B, we describe the adaptive partitioning of the spatial region and the spatial kernel mechanism used with the standard Hawkes process. In section III-C, we explain the procedure for estimating the times and locations of the samples using our model. In section III-D, we introduce the training algorithm for optimizing the model parameters and the objective function of our algorithm. In section IV, we present the experiment results on both simulated and real-life data. Finally, in section V, we give the concluding remarks.

\section{Model and Problem Description}

	We denote the matrices with boldface and uppercase letters, e.g. $ \textbf{X} $. $ \textbf{X}_{i, j} $ refers to the element of the matrix at the $ i $th row and the $ j $th column. $ \textbf{X}_{i,:} $ is the $ i $th row and $ \textbf{X}_{:,j} $ is the $ j $th column of the matrix $ \textbf{X} $. We denote the vectors with boldface lowercase letters, e.g. $ \textbf{x} = [x_0, x_1, ..., x_N] $ is a vector with length $ N $. The notation $ \textbf{x}^T $ refers to the ordinary transpose of a vector and $ \ell^2 $ norm of the vector $ \textbf{x} $ is $ ||\textbf{x}||^2 = \left\langle \textbf{x}, \textbf{x} \right\rangle = \textbf{x}^T\textbf{x} $. $ \odot $ operator is the element-wise multiplication and $ \mathbbm{1}_N $ is a column vector of ones with length $ N $. All vectors are column vectors and real vectors.

\begin{figure}
  \includegraphics[width=8.8cm]{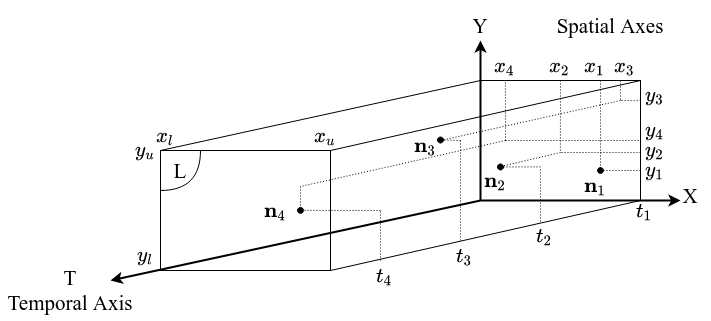}
  \caption{Illustration of a spatio-temporal sequence with 4 samples, $ \{\textbf{n}_1,\ \textbf{n}_2,\ \textbf{n}_3,\ \textbf{n}_4\} $. Each sample has a location stamp $ \{(x_1, y_1), (x_2, y_2), (x_3, y_3), (x_4, y_4)\} $ and a timestamp $ \{t_1, t_2, t_3, t_4\} $.}
  \label{fig:data_vis}
\end{figure}
	
	A spatio-temporal sequence consists of samples that are distributed along three axes: a temporal axis $ T $ and two spatial axes $ X $ and $ Y $. An example sequence of samples $ \textbf{N} = \{\textbf{n}_i\}_i $ are shown in Fig. \ref{fig:data_vis}. Each sample $ \textbf{n}_i = [t_i, \textbf{l}_i] $ corresponds to an event with its respective occurrence time $ t_i $ and its location $ \textbf{l}_i = (x_i, y_i) $, which is generally recorded as a latitude and longitude pair. All samples are observed in the spatial region $ L $ with boundaries $ L = [[x_\text{l}, x_\text{u}], [y_\text{l}, y_\text{u}]] $ as shown in Fig. \ref{fig:data_vis}, i.e. $ x_i \in [x_\text{l}, x_\text{u}] $, $ y_i \in [y_\text{l}, y_\text{u}]\ \forall \textbf{n}_i $. We model the data as a continuous sequence in both time and space, therefore $ t_i \in \Real $ and $ x, y \in \Real $. Moreover, samples are ordered according to their occurrence times, i.e., $ t_i > t_j $ for $ i>j $.
	
	We form the set $ \{t_i\}_{i} $, which consists of the sample observation times. Using the observation times set, we define the history $ \Omega(t) $ at time $ t $, which is expressed as $ \Omega(t) = \{\textbf{n}_i | t_i < t\}_{i} $, i.e. the observed samples until time $ t $. We aim to predict the observation time and location of future samples. We investigate this problem in two setups. In the first setup, we predict the number of events in a future spatio-temporal interval. This setup is preferable when the events are densely populated in small spatio-temporal intervals, such as the criminal activities in a city. We measure the performance for this setup via the squared error between the predicted number of events and the actual number of events in a fixed horizon. In the second setup, we predict the exact time and location of the next event $ \textbf{n}_i = [t_i, \textbf{l}_i] $ using the history $ \Omega(t_i) $. This setup is used for event sequences where occurrence times are distantly spaced in time, such as the earthquake events in an area. We assess the performance of our estimation $ \hat{\textbf{n}}_i $ for the sample $ \textbf{n}_i $ with the loss function
	
    \begin{equation}
    	\label{eqn:performance_metric_single}
    	l(\hat{\textbf{n}}_i, \textbf{n}_i) = [(\hat{t}_i-t_i)^2, (\hat{x}_i-x_i)^2 + (\hat{y}_i- y_i)^2].
    \end{equation}

    We model the sample observation times and locations with a point process model, therefore the number of events in an interval is computed by integrating the intensity function. The next event time and location are the expected time and location of the sample $ \textbf{n}_i $, i.e. $ (\hat{t}_i,\ \hat{\textbf{l}}_i) = \E[\tilde{t}_i,\ \tilde{\textbf{l}}_i | \Omega(t_{i-1})] $. Variables $ \tilde{t}_i $ and $ \tilde{\textbf{l}}_i $ are random variables, which are the observation time and spatial location of the sample $ \textbf{n}_i $. Therefore, we give brief information about the point processes in the next subsection.

\subsection {Representation of Sample Observations with Point Processes}

	We define the function $ N(t) $ as the total number of samples up to time $ t $. For point processes, the density function of the observation time of the sample $ \textbf{n}_i $ is given as
	
    \begin{equation}
    	\label{eqn:probability_density_function}
    	f_{\tilde{t}_i}(t | \Omega(t)) = \lim_{\Delta t\rightarrow 0} \text{P}(N(t) - N(t-\Delta t) = 1 | \Omega(t)),
    \end{equation}	
    which is a function of previous samples \cite{daley2007introduction}. We can also express the density function in \eqref{eqn:probability_density_function} using the conditional intensity function
    	
    \begin{equation}
    	\label{eqn:density_in_intensity}
    	f_{\tilde{t}_i}(t | \Omega(t)) = \lambda(t | \Omega(t)) e^{-\int_{t_{i-1}}^t \lambda(t' | \Omega(t')) dt'},
    \end{equation}		
     by approximating it with Bernoulli trials as $ \Delta t \rightarrow 0 $. Point generation mechanisms and the form of the probability density functions of a point process is controlled by the choice of the intensity function $ \lambda(t | \Omega(t)) $ \cite{daley2007introduction}. Definition of the intensity function is given as
    	
    \begin{equation}
    	\label{eqn:conditional_intensity_function}
    	\lambda(t | \Omega(t)) = \lim_{\Delta t \rightarrow 0} \frac{\text{P}(N(t+\Delta t) - N(t) = 1 | \Omega(t))}{\Delta t},
    \end{equation}	
    which corresponds to the expected number of observations in an infinitesimal time interval around $ t $. Thus, the expected number of events that will occur in a temporal interval is computed by integrating the intensity function over it.

	Different choices of the conditional intensity function will result in different sample generation mechanisms as the magnitude of the intensity control the rate at which the samples are observed. Depending on the formulation, this rate can be constant, time-varying or space-varying \cite{daley2007introduction}. We formulate our algorithm using the Hawkes process intensity, which is a function of the history as
	
    \begin{equation}
    	\label{eqn:hawkes_intensity}
    	\lambda(t | \Omega(t)) = \mu + \sum_{t_j<t} e^{-\gamma(t-t_j)},
    \end{equation}	
    where $ \{t_j\}_j $ are the times of the sample observations before $ t $, $ \mu $ is the background intensity and  $ \gamma $ is called the decay rate.

    \begin{remark}
    	Although we give our formulations for the Hawkes process intensity, our approach can be used with other point process intensity formulations. Thus, depending on the application, instead of using the Hawkes process, we could use any other intensity formulation such as the Poisson intensity
    	
    	\begin{equation}
    		\lambda(t | \Omega(t)) = \lambda^*,
    	\end{equation}
    	where $ \lambda^* \in \Real^+ $ \cite{cox1980point}. We could also use the intensity of the self-correcting process
    	
    	\begin{equation}
    		\lambda(t | \Omega(t)) = \mu t - \sum_{t_j < t} \alpha,
    	\end{equation}	 
    	$ \mu,\alpha \in \Real^+ $ \cite{isham1979self}.
    	
    \end{remark}

	We choose the Hawkes process intensity formulation due to its certain properties. First of all, the dependency on the past samples yields a non-stationary process in time due to the temporal kernel $ g_\text{t}(t-t_j) = e^{-\gamma(t-t_j)} $. Moreover, the effect of the past samples in the intensity is additive and increases with closer sample times. Thus, it is a self-exciting process. This property is useful for modeling certain real-life applications \cite{mohler2011self}, \cite{wang2018graph}. Nevertheless, spatial location information of the past samples are not incorporated into the formulation. We aim to predict the times and locations of the samples using the observation times and locations of the past samples. To this end, we introduce a conditional intensity function by incorporating the spatial interactions with a spatial kernel mechanism. In the following section, we give details about this spatial kernel.

\section{A Novel Spatio-temporal Prediction Model Based On Point Processes}
	Here, we introduce our formulation for partitioning a spatial region into fixed number of adaptive subregions using decision trees. We first introduce the adaptive decision tree structure and then describe the point process optimization procedure in each subregion to represent the sample observations. Finally, we introduce our training procedure and the likelihood-based objective function.

\subsection {Spatio-temporal Conditional Intensity Function with Spatial and Temporal Kernels}\

	Here, we give the formulation for the conditional intensity function that uses a temporal and spatial kernel to model the sample observations with a time and space-varying process. Our spatial kernel is based on decision trees, which forms adaptive spatial partitions and can be jointly optimized with the process parameters.
	
	Hawkes process intensity is a time-varying function and can change over time based on the past sample observations. However, in spatio-temporal data, the intensity could change in space as well. Thus, the intensity can be expressed as a function of both time and space to model non-stationary real-life data. Consequently, we define the intensity in \eqref{eqn:conditional_intensity_function} as the expected number of sample observations in an infinitesimal spatio-temporal interval around time $ t $ and location $ \textbf{l} $, which is given as
	
    \begin{equation}
    	\label{eqn:extended_conditional_intensity_function}
    	\lambda(t, \textbf{l} | \Omega(t)) = \lim_{\Delta t \rightarrow 0,\ \Delta l \rightarrow 0} \frac{\text{P}(N(t+\Delta t, \bar{\textbf{l}}) - N(t) = 1 | \Omega(t))}{\Delta t \Delta l}
    \end{equation}	
    where $ \bar{\textbf{l}} $ is the infinitesimal circular region centered at the location $ \textbf{l} $ with radius $ \Delta l $, i.e. $ \bar{\textbf{l}} = \{(x,\ y) \in \Real^2 |\ ||\textbf{l}-(x,\ y)||^2 \leq \Delta l\}$. Consequently, we define $ N(t, \bar{\textbf{l}}) $ as the number of events up to time $ t $ and the number of events around the spatial location $ \textbf{l} $ at time $ t $.

	Since we model the data with a time and space-varying process, we partition the spatial region $ L $ into fixed number of subregions, $ \{L_k\}_{k=1}^K $. These subregions form a partition of the whole space, i.e. $ \bigcup_{k=1}^K L_k = L $ and $ L_i \cap L_j = \emptyset $. We group the samples into subsets depending on their spatial locations, i.e. $ \textbf{N}_k=\{\textbf{n}_i|\textbf{l}_i\in L_k\}_i $, which also partition the samples. We also define the vector $ \boldsymbol{\rho}(\textbf{l}) = [\rho_k(\textbf{l})]_{k=1}^K $ $ \forall \textbf{l} \in L $, which is the subregion vector where 
	
    \begin{equation}
    	\label{eqn:subregion_vector}
    	\rho_k(\textbf{l})= 
    \begin{cases}
        1,& \text{if } \textbf{l} \in L_k\\
        0,              & \text{otherwise.}
    \end{cases}
    \end{equation}
    One-hot vector $ \boldsymbol{\rho}(\textbf{l}) $ indicates the subregion containing the location $ \textbf{l} $. For each subregion, we represent the sample observations with an individual point process model, hence with an individual conditional intensity. For the subregion $ L_k $, the conditional intensity is $ \lambda_k(t, \Omega(t)) $. As a result, we express the conditional intensity for any arbitrary location as $ \lambda(t, \textbf{l} | \Omega(t)) $, which is the conditional intensity of the subregion $ L_k $. It is given as,

    \begin{equation}
    	\label{eqn:spatio_temporal_intensity}
    	\lambda(t, \textbf{l} | \Omega(t))= 
    \begin{cases}
        \lambda_1(t | \Omega(t)),& \text{if } \textbf{l} \in L_1\\
        \lambda_2(t | \Omega(t)),& \text{if } \textbf{l} \in L_2\\
        ...\\
        \lambda_K(t | \Omega(t)),& \text{if } \textbf{l} \in L_K.\\
    \end{cases}
    \end{equation}
    
    We can also express the conditional intensity as
    	
    \begin{equation}
    	\label{eqn:prod_intensity}	
    	\lambda(t, \textbf{l} | \Omega(t)) = \boldsymbol{\rho}(\textbf{l})^T\boldsymbol{\lambda}(t, \textbf{l} | \Omega(t)),
    \end{equation}
    where $ \boldsymbol{\lambda}(t, \textbf{l} | \Omega(t)) = [\lambda_k(t, \textbf{l} | \Omega(t))]_{k=1}^K $. 
    	
	In order to incorporate the location information of the past samples, we also add a spatial kernel to the intensity $ \lambda_k(t | \Omega(t)) $ as 
	
    \begin{equation}
    	\label{eqn:spatial_kernel_hawkes}
    	\lambda_k(t | \Omega(t)) = \mu_k + \sum_{t_j<t} g_{\text{t},k}(t-t_j)g_{\text{l},k}(\textbf{l}_j),
    \end{equation}		
    where the temporal kernel $ g_{\text{t},k}(t-t_j) $ is selected as in \eqref{eqn:hawkes_intensity} due to the self-exciting property. We formulate the spatial kernel $ g_{\text{l},k}(\textbf{l}_j) $ in \eqref{eqn:spatial_kernel_hawkes} as
    	
    \begin{equation}
    	\label{eqn:spatial_kernel}
    	g_{\text{l},k}(\textbf{l}_j) = \mathbf{\Gamma}_{:,k}^T\boldsymbol{\rho}(\textbf{l}_j),
    \end{equation}
    where $ \mathbf{\Gamma} $ is a $ K \times K $ matrix modeling the interaction among the samples in all subregions, which is referred to as the interaction matrix. The element $ \mathbf{\Gamma}_{k, l} $ corresponds to the effect of a sample $ \textbf{n} \in \textbf{N}_k $ in the subregion $ L_k $ to the intensity of the $ l $th subregion. 

	Note that the summation in \eqref{eqn:spatial_kernel_hawkes} accumulates the effects of all the past samples. In online setups, this summation would grow indefinitely, however, the exponential kernel in the temporal kernel allows us to truncate the summation by including only the $ \nu $ most recent elements. To this end, we ignore the rest of the samples and define the set $ \textbf{N}_{\Omega}(t, \nu) $, which contains the most recent samples in the interval $ [t-\nu, t] $.
	
	We form the spatial subregions using decision trees that are adaptively organized in time. Thus, as the intensity functions are organized with new samples, boundaries of the spatial subregions are also organized. In the following subsection, we give details about the adaptive tree structure.

\subsection{Adaptive Partitioning of the Spatial Region with Decision Trees}
	
\begin{figure}
      \includegraphics[width=8.8cm]{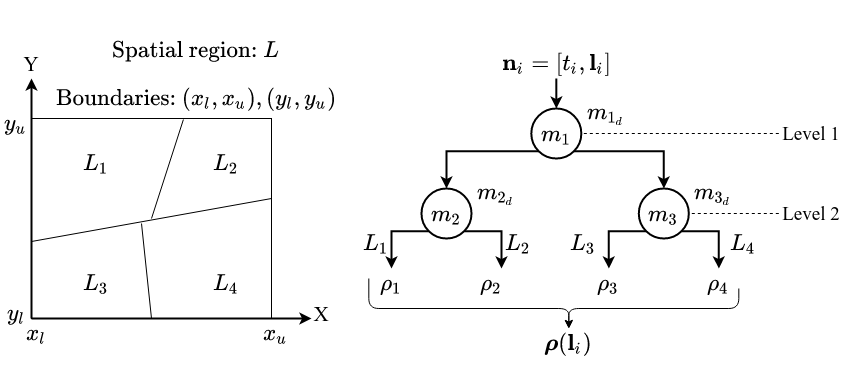}
      \caption{Diagram of a 2-level tree, which has three nodes $ m_1,\ m_2,\ m_3 $. At every node, a decision function is computed, which are shown as $ {m_1}_d,\ {m_2}_d,\ {m_3}_d $. Subregion scores $ \boldsymbol{\rho}(\textbf{l}_i) $ are computed for 4 branch leaves for the input samples $ \textbf{n}_i=[t_i,\textbf{l}_i] $. Each leaf node is associated with a spatial subregion, which are $ L_1 $, $ L_2 $, $ L_3 $ and $ L_4 $ for the 4 branches.}
      \label{fig:tree}
    \end{figure}

	The decision tree in our algorithm consists of a collection of nodes $ D = \{m_r\}_{r=1}^R $ that are placed hierarchically as in Fig. \ref{fig:tree}. Each node, except the leaf nodes (located at the bottom of the tree, at level $ \ell=2 $), has two children that are linked with branches. The top node, $ m_1 $, is referred to as the root node.

	The sample $ \textbf{n}_i $ is assigned to a spatial subregion using the decision tree. At each node, the sample is either assigned to the left or right node starting from the root node until reaching to a leaf node. A decision function, $ {m_r}_\text{d} $ with range $ \{-1, +1\} $, is computed at each node, which uses the features of the input sample. In certain works, a randomly selected feature from the input sample is used at each node for comparison \cite{ho1995random}, \cite{ali2012random}, where the input has features $ \textbf{n}_i = [n_i^{(f)}]_{f=1}^F $. The decision function is given as

    \begin{equation}
    	\label{eqn:decision_tree_simple}
    	{m_r}_\text{d}(\textbf{n}_i) = \sign (n_i^{({m_r}_\text{f})} - {m_r}_\text{b}) \lessgtr 0,
    \end{equation}	
    where $ n_i^{({m_r}_\text{f})} $ is the feature used at the node $ m_r $ and $ {m_r}_\text{b} $ is the threshold for comparison. The sample $ \textbf{n}_i $ is assigned to the left node if \eqref{eqn:decision_tree_simple} is positive and to the right node otherwise.

	Samples are separated by their features with the decision functions as in \eqref{eqn:decision_tree_simple}. In our case, features of the samples are only the location vector $ \textbf{l} = (x_i, y_i) $. Therefore, we separate the samples by comparing their spatial components with threshold values. As the depth of the tree grows, the number of spatial subregions will also increase. For a depth-$ \ell $ decision tree, we have $ K=2^\ell $ leaf branches for the subregions $ \{L_k\}_{k=1}^{K} $ as in Fig. \ref{fig:tree}.
	
	Instead of using a single feature for comparison as in \eqref{eqn:decision_tree_simple}, we use all features $ [n_i^{(f)}]_{f=1}^F $ for comparison. Hence, the decisions have the form
	
    \begin{equation}
    	\label{eqn:decision_tree_rotated}
    	{m_r}_\text{d}(\textbf{n}_i) = \sign ({m_r}_\textbf{w}^T\textbf{l}_i - {m_r}_\text{b}) \lessgtr 0,
    \end{equation}	
    where $ {m_r}_\textbf{w} $ is the weight vector combining the location elements $ x_i,\ y_i $. With such decision functions, we partition the space with lines as in Fig. \ref{fig:tree}. 

    \begin{remark}
	We use a decision tree to separate the feature space, which results in linear decision boundaries. We could use any nonlinear function of the spatial coordinates, e.g.
	
	\begin{equation}
		\textbf{f}(x_i, y_i) = g(\textbf{w}_x x_i + \textbf{w}_y y_i + b),
	\end{equation}
	where $ \{\textbf{w}_x, \textbf{w}_y\} $ are $ K\times1 $ weight vectors and $ g $ is a nonlinear function. We can generate the vector $ \boldsymbol{\rho}(x_i, y_i) $ by setting the maximum element of $ \textbf{f}(x_i, y_i) $ to 1 and the rest to 0, which generates a one-hot vector as in the decision tree partitioning. However, this formulation for the spatial separation could result in a complex partitioning \cite{csaji2001approximation}. 

    \end{remark}
	
	We optimize the boundaries of the spatial region by updating the weight parameters in \eqref{eqn:subregion_vector} via a gradient-based procedure. Instead of using hard functions in the decision functions, we use the sigmoid function ($ \sigma(x) = 1/(1+e^{-x}) $) as
	
    \begin{equation}
    	\label{eqn:decision_tree_soft}
    	{m_r}_\text{d}(\textbf{n}_i) = \sigma ({m_r}_\textbf{w}^T\textbf{n}_i - {m_r}_\text{b}),
    \end{equation}	
    which yields a score in $ [0, 1] $. Soft functions in decision functions as in \eqref{eqn:decision_tree_soft} were previously used in \cite{willems1995context} for source coding. In the standard formulation given in \eqref{eqn:decision_tree_rotated}, a sample is either assigned to the left branch or to the right branch. As a result, a sample can only lie in a single subregion, which was expressed with a one-hot vector as in \eqref{eqn:subregion_vector}. However, by using a soft function in the decision function, this hard separation is removed. In this case, the sample $ \textbf{n}_i $ is assigned to both the left branch and to the right branch with scores $ {m_r}_\text{d}(\textbf{n}_i) $ and $ 1-{m_r}_\text{d}(\textbf{n}_i) $.

	In the standard formulation with hard decision functions, we represent the spatial subregion score vector with a one-hot vector. However, in this case, the subregion score vector is $ \boldsymbol{\rho}(\textbf{l}_i) = [\rho_k(\textbf{l}_i)]_{k=1}^K $ where $ \rho_k(\textbf{l}_i) \in [0, 1]\ \forall k $. We compute the scores by multiplying the scores obtained at each layer starting from the root node until the leaf branches as
	
    \begin{equation}
    	\label{eqn:subregion_score}
    	\rho_k(\textbf{n}_i) = \prod_{l=1}^\ell \pi(\textbf{n}_i, \boldsymbol{\nu}_k^{(l)}, \boldsymbol{\tau}_k^{(l)}),
    \end{equation}	
    where 
    
    \begin{equation}
    	\label{eqn:subregion_score_level}
    	\pi(\textbf{n}_i, \boldsymbol{\nu}_k^{(l)}, \boldsymbol{\tau}_k^{(l)})= 
    \begin{cases}
        {\nu_k^{(l)}}_\text{d}(\textbf{n}_i),& \text{ } \tau_k^{(l)} = \text{left}\\
        1-{\nu_{k^{(l)}}}_\text{d}(\textbf{n}_i),& \text{ } \tau_k^{(l)} = \text{right}.\\
    \end{cases}
    \end{equation}	
    We define the sequence of branches and sequence of visited nodes to reach the $ k $th leaf as $ \boldsymbol{\tau}_k $ and $ \boldsymbol{\nu}_k $, e.g. for the depth-2 tree shown in Fig. \ref{fig:tree}, second leaf branch $ \rho_2 $ is reached by taking first the left branch and then the right branch. Thus, $ \boldsymbol{\tau}_2 = [\text{left, right}]$ and $ \boldsymbol{\nu}_2 = [m_1, m_2] $. $ \boldsymbol{\nu}_k^{(l)} $ in \eqref{eqn:subregion_score} and \eqref{eqn:subregion_score_level} is the visited node at level $ l $ to reach the $ k $th leaf. $ \boldsymbol{\tau}_k^{(l)} $ is the direction of the branch at level $ l $ to reach the $ k $th leaf. The decision function score at level $ l $ is expressed as $ \pi(\textbf{n}_i, \boldsymbol{\nu}_k^{(l)}, \boldsymbol{\tau}_k^{(l)}) $.
    
    \begin{remark}
    \label{tree_remark}
    	The following properties are satisfied by the subregion score vector:
    
    \begin{itemize}
    	\item The maximum subregion score is obtained for the subregion containing the spatial location, $ \textbf{l}_i $ of the sample $ \textbf{n}_i $, i.e. $ \kappa = \argmax_{k \in \{0, 1, ..., K\}} [\rho_k(\textbf{n}_i)] \Leftrightarrow \textbf{l}_i \in L_\kappa $.
    	
    	\item Subregion scores sum up to 1 for any location $ \textbf{l}_i \in L $, i.e. $ \sum_{k=1}^K \rho_k(\textbf{l}_i) = 1 $.
    \end{itemize}	
    
    \end{remark}
    
	The spatial subregion vector of scores $ \boldsymbol{\rho}(\textbf{n}_i) $ are computed by an adaptive decision tree, which is used in the spatial kernel mechanism in \eqref{eqn:spatial_kernel}.	In the next section, we explain the estimation method for the sample times and locations using the spatio-temporal density function. Moreover, we also explain the procedure for optimizing the model parameters.

\subsection {Sample Time and Location Prediction}

	In order to estimate the time and location of samples, we formulate the joint density function $ f_{\tilde{t_i}, \tilde{\textbf{l}}_i} $, using the intensity \eqref{eqn:extended_conditional_intensity_function} and the density in \eqref{eqn:density_in_intensity}. The joint density function is
	
    \begin{equation}
    \begin{split}
    	\label{eqn:joint_density_function}
    	f_{\tilde{t_i}, \tilde{\textbf{l}}_i}(t, \textbf{l} | \Omega(t_{i-1})) = \lambda(t, \textbf{l} | \Omega(t_{i-1}))e^{- \Lambda_{t_{i-1}}(t, \textbf{l}) }, \\
    	\Lambda_{t_{i-1}}(t, \textbf{l}) = \int_{t_{i-1}}^t\int_L \lambda(t', \textbf{l}' | \Omega(t_{i-1}))d\textbf{l}'dt'.
    \end{split}
    \end{equation}	
    We use a more compact expression for the conditional intensity as in \eqref{eqn:prod_intensity}, which corresponds to the weighted average of the subregion intensities $ \boldsymbol{\lambda}(t, \textbf{l} | \Omega(t_{i-1})) $. Therefore,
    	
    \begin{equation}
    	\label{eqn:smooth_intensity}
    	\lambda(t, \textbf{l} | \Omega(t_{i-1})) = \boldsymbol{\rho}(\textbf{l})^T\boldsymbol{\lambda}(t, \textbf{l} | \Omega(t_{i-1})).
    \end{equation}

	Note that in \eqref{eqn:smooth_intensity}, the computations include an integration. We estimate the integration with the Riemann sum \cite{flannery1992numerical}. Thus, uniformly spaced samples $ \{\textbf{n}_j\}_j $ in time and space are sampled for the computation. For example, the exponent $ \Lambda_{t_{i-1}}(t, \textbf{l}) $ is calculated as
	
    \begin{equation}
    	\label{eqn:riemann_exponent}
    	\Lambda_{t_{i-1}}(t, \textbf{l}) = \frac{1}{M} \sum_{t_j, \textbf{l}_j} \lambda(t_j, \textbf{l}_j | \Omega(t_{i-1}))|T||L|,
    \end{equation}
    where $ M $ is the number of points used in the integration estimation. In our experiments, we have observed that choosing a sufficiently large number of points does not cause any negative effects on the performance. We sample the points $ \textbf{n}_j $ at times $ t_j $ and at locations $ \textbf{l}_j $. Finally, the terms $ |T| $ and $ |L| $ are the length and area of the temporal and spatial boundaries of the integration respectively. We compute $ |T| = t-t_{i-1} $ and $ |L| = (x_\text{u}-x_\text{l})\times (y_\text{u}-y_\text{l}) $. 
    
    \begin{remark}
    \label{monte_carlo}
    	Instead of using the Riemann sum for the integration, we could use the Monte Carlo Integration method \cite{caflisch1998monte}, which is preferred over Riemann sum when the function has abrupt changes. To circumvent this issue, we can readily increase the number of sampled points in the Riemann sum given in \eqref{eqn:riemann_exponent}. In our experiments, we have observed that choosing a sufficiently large number of samples does not have a negative effect on the performance. 
    
    \end{remark}
    
    	To estimate the sample time and location, we marginalize the joint density function over time and space to obtain the marginal density functions as
    	
    \begin{equation}
    \begin{split}
    	\label{eqn:marginal_pdfs}
    	f_{\tilde{t}_i}(t) &= \int_L f_{\tilde{t_i}, \tilde{\textbf{l}}_i}(t, \textbf{l} | \Omega(t_{i-1}))d\textbf{l},\\
    	f_{\tilde{\textbf{l}}_i}(\textbf{l}) &= \int_{t_{i-1}}^\infty f_{\tilde{t_i}, \tilde{\textbf{l}}_i}(t, \textbf{l} | \Omega(t_{i-1}))dt.
    \end{split}
    \end{equation}
    We compute the estimated time and location of the sample $ \textbf{n}_i $ as the conditional mean of the random variables $ \tilde{t_i} $ and $ \tilde{\textbf{l}_i} $ as
    	
    \begin{equation}
    	\label{eqn:mean_estimate_t}
    	\hat{t}_i = \E [\tilde{t}_i | \Omega(t_{i-1})] = \int_{t_{i-1}}^\infty t'f_{\tilde{t}_i}(t' | \Omega(t_{i-1}))dt',
    \end{equation}	
    
    \begin{equation}
    	\label{eqn:mean_estimate_m}
    	\hat{x}_i = \E [\tilde{x}_i | \Omega(t_{i-1})] = \int_L x'f_{\tilde{\textbf{l}}_i}(m'| \Omega(t_{i-1}))dl,
    \end{equation}	
    
    \begin{equation}
    	\label{eqn:mean_estimate_n}
    	\hat{y}_i = \E [\tilde{y}_i | \Omega(t_{i-1})] = \int_L y'f_{\tilde{\textbf{l}}_i}(n'| \Omega(t_{i-1}))dl,
    \end{equation}	
    where the integration in \eqref{eqn:mean_estimate_m} and \eqref{eqn:mean_estimate_n} is over the region $ L $.
	
	The integration computations in \eqref{eqn:marginal_pdfs} and \eqref{eqn:mean_estimate_t} are over an infinitely long time interval. However, the conditional intensity function in \eqref{eqn:conditional_intensity_function} is, by definition, a non-negative function, and thus \eqref{eqn:joint_density_function} is a monotonically decreasing function of time. Hence, we compute the Riemann sum of the integration up to a certain time. We experiment with different values for this parameter.
	
	The integration computation in \eqref{eqn:joint_density_function} is between the time of the last event and any time $ t $. Computing this integration for two arbitrary times $ t' > t'' > t_{i-1} $ from scratch would be unnecessary as they share a common interval. Therefore, we sort the sampled points in the Riemann sum estimation in \eqref{eqn:riemann_exponent}. Starting from the closest point in time, we accumulate the intensity towards the furthest point and keep the accumulated intensity in memory.

\subsection {Model Parameter Optimization via Likelihood Maximization}	
	
	Here, we describe the objective function that we are using to update the model parameters and the gradient based optimization procedure.
	
	The parameters of the node $ m_r $ are the weights $ {m_r}_\textbf{w} $ and the threshold $ {m_r}_\text{b} $. For a decision tree with $ R $ nodes, the model parameters are the collection of node parameters $ \boldsymbol{\Theta}_{\text{tree}} = \{({m_r}_\textbf{w}, {m_r}_\text{b})\}_{r=1}^R $. We can also group the point process parameters for $ K $ subregions. For a single subregion, the intensity in \eqref{eqn:spatial_kernel_hawkes} has the parameters $ \mu_k $ and $ \gamma_k $. Interaction matrix $ \mathbf{\Gamma} $ is a common parameter for all subregions. All parameters for all subregions are $ \boldsymbol{\Theta}_\text{hawkes} = \{\mathbf{\Gamma},\ \{(\mu_k, \gamma_k)\}_{k=1}^K\} $.
	
	Combining the two sets, we form the set of model parameters $ \Theta = [\Theta_\text{tree}, \Theta_\text{hawkes}] $. Full notation for the density function $ f_{\tilde{t_i}, \tilde{\textbf{l}}_i}(t, \textbf{l} | \Omega(t_{i})) $ and $ \lambda(t_j, \textbf{l}_j | \Omega(t_{j})) $ are $ f_{\tilde{t_i}, \tilde{\textbf{l}}_i}(t, \textbf{l} | \Omega(t_{i}), \Theta) $ and $ \lambda(t_j, \textbf{l}_j | \Omega(t_{i}), \Theta) $ respectively. We drop the term $ \Theta $ from the notation for simplicity.
	
	We compute the likelihood of our model using
	
    \begin{equation}
    	\label{eqn:likelihood}
    	\Lagr (\textbf{N}) = \prod_{i=1}^I f_{\tilde{t_i}, \tilde{\textbf{l}}_i}(t_i, \textbf{l}_i | \Omega(t_{i-1})).
    \end{equation}
	We find the optimal set of model parameters $ \boldsymbol{\Theta}^* $ by randomly initializing a set of model parameters $ \boldsymbol{\Theta}_0 $ and updating the set with the stochastic gradient-ascent algorithm \cite{bottou2010large}. To remove the exponential terms in \eqref{eqn:likelihood}, we compute the gradients with respect to the log-likelihood $ \tilde{\Lagr} = \log \Lagr $, which is

    \begin{equation}
    	\label{eqn:log_likelihood}
    	\tilde{\Lagr} (\textbf{N}) = \sum_{i=1}^I \log f_{\tilde{t_i}, \tilde{\textbf{l}}_i}(t_i, \textbf{l}_i | \Omega(t_{i-1})).
    \end{equation}	
    Using the definition of $ f_{\tilde{t_i}, \tilde{\textbf{l}}_i}(t_i, \textbf{l}_i | \Omega(t_{i-1})) $ in \eqref{eqn:log_likelihood} with the formulation in \eqref{eqn:joint_density_function}, we obtain
    	
    \begin{equation}
    	\label{eqn:expanded_log_likelihood}
    	\tilde{\Lagr} (\textbf{N}) = \sum_{i=1}^I \log \lambda(t_i, \textbf{l}_i | \Omega(t_{i-1})) - \sum_{j=1}^{J} \Lambda_{t_{i-1}}(t_j, \textbf{l}_j),
    \end{equation}		
    which has two terms, $ \Lagr_\text{positive} $ and $ \Lagr_\text{negative} $. The integration in $ \Lagr_\text{negative} $ can be accumulated in time and expressed as a single integration as
    	
    \begin{equation}
    	\label{eqn:log_negative}
    	\Lagr_\text{negative} = \int_{t_{0}}^T\int_L \lambda(t', \textbf{l}' | \Omega(t_{i-1}))d\textbf{l}'dt',
    \end{equation}	
    where $ t_0 $ is the start of the integration simulation and $ T $ is the end time. We sample uniformly spaced points in both time and space for the integration in \eqref{eqn:log_negative} as in \cite{mei2017neural}. $ \Lagr_\text{positive} $ accumulates the log-intensities of the sample observations and $ \Lagr_\text{negative} $ penalizes the intensity function for the sampled points, which represent the rest of the spatio-temporal interval. We sum the two terms with a weight term as
	
    \begin{equation}
    	\label{eqn:weighted_objective}
    	\tilde{\Lagr} = \Lagr_\text{positive} + \alpha \Lagr_\text{negative},
    \end{equation}	
    to control the effect of the negative and the positive terms on the overall $ \tilde{\Lagr} $.
	
	We define the optimal parameters as the set of model parameters $ \boldsymbol{\Theta}^* $ that exceeds a certain tolerance level for the log-likelihood $ \Lagr_\text{tol} $, i.e.
	
    \begin{equation}
    	\label{eqn:likelihood_condition}
    	\Lagr (\textbf{N} | \Theta^*) \geq \Lagr_\text{tol}
    \end{equation}
    To compute the optimal set of parameters, we split the given set of samples $ \textbf{N} = \{\textbf{n}_i\}_{i=1}^I $ into three subgroups; training set ($ \textbf{N}_\text{train} = \{\textbf{n}_i\}_{i=1}^{I_\text{train}} $), validation set ($ \textbf{N}_\text{val} = \{\textbf{n}_i\}_{i=I_\text{train}}^{I_\text{val}} $) and test set ($ \textbf{N}_\text{test} = \{\textbf{n}_i\}_{i=I_\text{val}}^{I} $). We use the training set for the parameter optimization.
	
	We update the set of parameters starting from the initial values $ \boldsymbol{\Theta}_0 $ with the gradient updates. For a model parameter $ \boldsymbol{\theta} \in \boldsymbol{\Theta} $ we have
	
    \begin{equation}
    	\label{eqn:first_order_grad}
    	\boldsymbol{\theta}_{i+1} = \boldsymbol{\theta}_i + \eta \nabla_{\boldsymbol{\theta}} \tilde{\Lagr}(\textbf{N}_B),
    \end{equation}	
    where $ \textbf{N}_B $ is a mini-batch consisting of randomly picked samples from the training set $ \textbf{N}_\text{train} $. The learning rate $ \eta $ scales the magnitudes of parameter updates and $ \nabla_{\boldsymbol{\theta}} \tilde{\Lagr}(\textbf{N}_b) $ is the gradient vector of $ \tilde{\Lagr}(\textbf{N}_b) $ with respect to the parameter $ \boldsymbol{\theta} $. Using \eqref{eqn:expanded_log_likelihood} and \eqref{eqn:weighted_objective}, we have

    \begin{equation}
    	\label{eqn:lambda_derivative}
    	\frac{\partial \tilde{\Lagr}(\textbf{N}_b)}{\partial \boldsymbol{\theta}} = \frac{1}{B}\sum_{b=1}^B \left(\left[\frac{1}{\lambda_k (t_b, \textbf{l}_b)}\right]_{k=1}^K + \alpha \mathbbm{1}_K\right)\odot \frac{\partial \boldsymbol{\lambda}(t_b, \textbf{l}_b)}{\partial \boldsymbol{\theta}}.
    \end{equation}	
    Partial derivatives of $ \boldsymbol{\lambda} $ with respect to the Hawkes parameters are

    \begin{equation}
    	\label{eqn:del_derivative}
    	\frac{\partial \mathbf{\lambda}_k(t_b, \textbf{l}_b)}{\partial \gamma_k} = \mathbf{\Gamma}_{:, k}^T \sum_{\textbf{n} \in \textbf{N}_{\Omega}(t_b, \nu)} \rho_k(\textbf{l})(t - t_b)e^{\gamma_k (t-t_b)},
    \end{equation}
    
    \begin{equation}
    	\label{eqn:gamma_derivative}
    	\frac{\partial \mathbf{\lambda}_k(t_b, \textbf{l}_b)}{\partial \mathbf{\Gamma}_{i, k}} =  \sum_{\textbf{n} \in \textbf{N}(t_b, \nu)} \rho_i(\textbf{l}) e^{\gamma_{k}(t-t_{b})}
    \end{equation}
    and
    
    \begin{equation}
    	\label{eqn:mu_derivative}
    	\frac{\partial \lambda_k(t_b, \textbf{l}_b)}{\partial \mathbf{\mu}_k} =  \rho_k(\textbf{l}_b).
    \end{equation}	
    Similarly, the derivatives of the tree parameters are
    
    \begin{equation}
    	\label{eqn:p_derivative}
    	\frac{\partial \boldsymbol{\lambda}_k(t_b, \textbf{l}_b)}{\partial \boldsymbol{\rho}} =  \boldsymbol{\lambda}(t_b, \textbf{l}_b)\odot \left(\mathbbm{1}_K + \textbf{J}^T\textbf{p}(\textbf{l}_b) \right),
    \end{equation}	
    where $ \textbf{J} $ is the jacobian matrix. The element $ \textbf{J}_{i, j} $ corresponds to $ \partial \mathbf{\lambda}_i / \partial \rho_j $. Derivative of the weight vector $ {m_r}_\textbf{w} $ is
    
    \begin{equation}
    	\label{eqn:w_derivative}
    	\frac{\partial \boldsymbol{\rho}(\textbf{l}_b)}{\partial {m_r}_\textbf{w}} =  \sum_{k=1}^K \prod_{\substack{l=1,\\ l\neq l'}}^L \pi(\textbf{n}_b, \nu_k^{(l)}, \tau_k^{(l)})\sigma(1-\sigma)\textbf{l}_b.
    \end{equation}	
    
    where $ l' $ is the level of the node $ m_r $. Similarly,
    
    \begin{equation}
    	\label{eqn:b_derivative}
    	\frac{\partial \boldsymbol{\rho}(\textbf{l}_b)}{\partial {m_r}_\textbf{b}} =  \sum_{k=1}^K \prod_{\substack{l=1, \\l\neq l'}}^L \pi(\textbf{n}_b, \nu_k^{(l)}, \tau_k^{(l)})\sigma(1-\sigma),
    \end{equation}	
    where $ \sigma = \sigma({m_r}_\textbf{w}^T\textbf{l}_b + {m_r}_\text{b}) $.

	By the definition in \eqref{eqn:extended_conditional_intensity_function}, the conditional intensity should be a non-negative function. Optimizing the model parameters with gradient based updates without imposing any constraint could result in a negative intensity. Thus, we use the softplus function

    \begin{equation}
    	\label{eqn:soft_intensity}
    	\tilde{\lambda}(t, \textbf{l}|\Omega(t)) = \log (1+e^{\lambda(t, \textbf{l}|\Omega(t))})
    \end{equation}	
    to make the intensity non-negative through the optimization steps.

    \begin{algorithm}
    \caption{Likelihood based first-order gradient training procedure.}
    \label{alg:grad_algo}
    
    \begin{algorithmic} 
    \REQUIRE Hyperparameters, Samples $ \textbf{N} $
    
    \STATE $ i=0 $
    \STATE Randomly initialize parameters, $ \boldsymbol{\theta}_\text{tree} $, $ \boldsymbol{\theta}_\text{hawkes} $.
    
    \STATE $ \boldsymbol{\Theta}_i = [\boldsymbol{\theta}_{tree}, \boldsymbol{\theta}_\text{hawkes}] $
    \STATE Sample $ J $ negative points $ \{[t_j, x_j, y_j]\}_{j=1}^J $
    \STATE $ \tilde{\Lagr}_\text{positive} \leftarrow \sum_{i=1}^I \log \lambda(t_i, \textbf{l}_i|\Omega(t_{i-1})) $
    \STATE $ \tilde{\Lagr}_\text{negative} \leftarrow \sum_{j=1}^J \Lambda_{t_j}(t_j, \textbf{l}_j) $
    \STATE $ \tilde{\Lagr} \leftarrow \tilde{\Lagr}_\text{positive} + \alpha \tilde{\Lagr}_\text{negative} $
    \WHILE{$ \tilde{\Lagr} < \log \Lagr_\text{tol} $} 
    	\STATE Sample $ J $ negative points $ \textbf{J} \leftarrow \{[t_j, x_j, y_j]\}_{j=1}^J $
    	\STATE $ \tilde{\Lagr}_\text{positive} \leftarrow \sum_{i=1}^I \log \lambda(t_i, \textbf{l}_i|\Omega(t_{i-1})) $
    	\STATE $ \tilde{\Lagr}_\text{negative} \leftarrow \sum_{j=1}^J \Lambda_{t_j}(t_j, \textbf{l}_j) $
    	\STATE $ \tilde{\Lagr} \leftarrow \tilde{\Lagr}_\text{positive} + \alpha \tilde{\Lagr}_\text{negative} $
    	\STATE Compute updates, $ \Delta \boldsymbol{\theta}_i \leftarrow \nabla_{\boldsymbol{\theta}} \tilde{\Lagr}(\textbf{N}, \textbf{J}) $
    	\STATE $ \boldsymbol{\theta}_{i+1} \leftarrow \boldsymbol{\theta}_i + \eta*\Delta \boldsymbol{\theta}_i $
    	\STATE $ i \leftarrow i+1 $
    \ENDWHILE
    \end{algorithmic} 
    
    \end{algorithm}
	
	Our optimization algorithm is presented in Algorithm \ref{alg:grad_algo}, which is an iterative optimization procedure. We first randomly initialize a decision tree with depth $ \ell $ and point processes corresponding to all subregions. At the $ i $th iteration, we first compare the current log-likelihood with the tolerance level $ \Lagr_\text{tol} $. If it is not exceeded, we compute the positive log-likelihood of the model with the current parameters. We also sample points in the spatio-temporal interval and compute the negative term in the log-likelihood using these points. Using the log-likelihood, we compute the parameter updates, $ \Delta \boldsymbol{\theta}_i $ for each model parameter. If the tolerance level is achieved at a step, we terminate the training process.
	
	To find the best set of hyperparameters, we perform multiple training runs with a different set of hyperparameters. At the end of each run, we compute the validation performance on the validation set. We pick the set that yields the highest log-likelihood. After the best set of hyperparameters are found, we compute the test performance with the same metric.
	
	In our training procedures, we use the stochastic gradient-ascent method, which uses the first order derivatives to update the model parameters with respect to an objective function. To increase the convergence speed of our training procedures, we also use the ADAM optimizer \cite{kingma2014adam}, which uses the first order derivatives. Thus, it does not introduce any computational cost for the update equations.


\section{Experiments}
    In this section, we present the experiment setup and the result comparisons on different datasets and for different models. We first describe the real-life datasets by giving brief information about their content and how they are processed. Then, we explain the algorithms that we use in our comparison. Finally, we investigate the effect of choosing different hyperparameters in our model by giving empirical results. 

    As described earlier, we assess the performance of our model in two setups. For event count estimation in a window, we use a real-life criminal events dataset and an earthquake events dataset. For the precise prediction of event times and locations, we use only the earthquake dataset. Details for each setup are given in the following sections.

    \subsection{Real-life Datasets For Spatio-Temporal Prediction}
        Here, we explain the datasets used in the comparison of our algorithm with the existing approaches. The first dataset we use is the Chicago Crime Dataset \cite{department_2021}. The dataset contains criminal event records that occurred in Chicago City between 2001 and 2021. Each record is tagged with its date, latitude, and longitude. Also, each event has a description of the event details and an event class. We focus our attention on the recent entries between 2012 and 2017. Moreover, we only consider the time and location-stamps of the events for predicting future events.
        
        Since many criminal activities take place even in small spatial and temporal subintervals, rather than predicting the individual event time and locations, we estimate the total number of events that will occur in a specified temporal horizon and spatial area. The details of this evaluation are given in the following subsection.
        
        The second dataset is the Significant Earthquakes dataset \cite{EarthquakesDataset}. The dataset is recorded by the National Earthquake Information Center (NEIC). The dataset contains entries of earthquake event records with their latitude, longitude, depth, magnitude and date. It covers all the earthquake events from 1900 to 2000 that have a magnitude higher than 5.5. Location and time prediction of the earthquake events using the past data fits our problem description since the events are spatio-temporal samples.
	
    	Since the dataset only contains the earthquake event records with magnitudes higher than 5.5, it excludes the aftershock records. Aftershocks of earthquakes could have been effectively modeled by the Hawkes process formulation \cite{kanamori2003earthquake}. To this end, we choose a particular spatial region $ L $ between the latitudes 31.92$^\circ$ and 72.05$^\circ$ and between the longitudes 110.2$^\circ$ and 180.1$^\circ$ from the whole data, which contains a large number of sequential earthquakes. This region corresponds to the east of China and Russia, and all of Japan and Korea.
    	
    	We convert the earthquake records to spatio-temporal sequence $ \textbf{N} = \{\textbf{n}_i\}_{i=1}^{I} $. In particular, we generate 20 sequences by splitting the data into equal-length parts in time, i.e. 5 years. We further process the data by converting the dates of the events to time differences with respect to the first event time in seconds and scale the differences as $ \{t_i/(3600\times 24\times 30)\}_{i=1}^I $. Moreover, the spatial locations, i.e. latitude and longitude of the events are scaled as $ L = \{(x,\ y)\in \Real^2 | -10<x<10,\ -10<y<10\} $.
    	
    	Since the earthquake events are separated distantly in time, we estimate the time and location of specific events directly. Thus, we assess the performance of our algorithm directly on these estimates, which we explain in the following subsection.
    	
    	Finally, we also measure the performance of our algorithm on a simulated dataset. This dataset is artificially generated via the Thinning algorithm \cite{ogata1981lewis} using the same model structure. 

    \subsection{Benchmark Models For Performance Comparison}
        In this section, we describe the algorithms that we use in our benchmark for performance comparison. Since we investigate the performance of our algorithm on estimating both the total number of events and the event time and locations, we use two different set of algorithms.
        
        \subsubsection{Event Count Estimation Models}
            We estimate the total number of event in the horizon window by accumulating the intensity function. Although our model can estimate the number of events in any arbitrary interval, for evaluation purposes, we discretize the spatial region into grids with a selected resolution. Consequently, we obtain a 2-D array containing the number of events in each grid. Hence, we generate a 2-D frame containing the expected number of events in the prediction horizon for the spatial region.
            
            Finally, since the real-life data consists of exact event dates, latitudes and longitudes, we convert these values to relative position stamps with selected temporal and spatial resolutions. Moreover, we assess the performance of the predictions using the Root Mean-Squared-Error (RMSE) metric for each grid. 
    	
        	We use a CNN model, which processes the discrete data structure and predicts the event amount in a specified horizon. For every time step, we make a prediction using the most recent frames. The model applies several convolutional kernels to the most recent frames and generates an array containing the counts.
        	
        	During the training, we minimize the mean squared error between the generated array and the ground truth array over the whole training session. We update the model parameters using the ADAM optimizer with respect to the mean squared error between the generated frames and the ground-truth frames.
        
        \subsubsection{Event Time and Location Prediction Models}
            The first model in this comparison is the linear regression (we refer to as Linear), which predicts the displacement in time and space using the past displacement vectors with a linear model. Predictions are computed as
        	
        	\begin{equation}
        		\begin{split}
        			&\Delta t_i = \textbf{w}_{\text{tt}}^T [\Delta t_{i-k}]_{k=1}^K + \textbf{w}_{\text{tx}}^T [\Delta x_{i-k}]_{k=1}^K + \textbf{w}_{\text{ty}}^T [\Delta y_{i-k}]_{k=1}^K, \\
        			&\Delta x_i = \textbf{w}_{\text{xt}}^T [\Delta t_{i-k}]_{k=1}^K + \textbf{w}_{\text{xx}}^T [\Delta x_{i-k}]_{k=1}^K + \textbf{w}_{\text{xy}}^T [\Delta y_{i-k}]_{k=1}^K, \\
        			&\Delta y_i = \textbf{w}_{\text{yt}}^T [\Delta t_{i-k}]_{k=1}^K + \textbf{w}_{\text{yx}}^T [\Delta x_{i-k}]_{k=1}^K + \textbf{w}_{\text{yy}}^T [\Delta y_{i-k}]_{k=1}^K,
        		\end{split}
        	\end{equation}
        	which are using the $ K $ most recent observations.
        	
        	Another method for comparison is the standard RNN model, which predicts the displacement in time and space using the state vector. The state transition is computed as
        	
        	\begin{equation*}
        		\textbf{h}_i = \text{tanh}(\textbf{W}_\text{x}^T[\Delta t_{i-1}, \Delta x_{i-1}, \Delta y_{i-1}] + \textbf{W}_\text{h}^T\textbf{h}_{i-1}),
        	\end{equation*}
        	where $ \text{tanh} = (e^x-e^{-x})/(e^x+e^{-x}) $. We compute the displacements as
        	
        	\begin{equation*}
        		\begin{split}
        			&\Delta t_i = \textbf{w}_\text{to}^T \textbf{h}_i,\\
        			&\Delta x_i = \textbf{w}_\text{xo}^T \textbf{h}_i,\\
        			&\Delta y_i = \textbf{w}_\text{yo}^T \textbf{h}_i.
        		\end{split}
        	\end{equation*}
        	We also repeat the same analysis for the LSTM model.
        	
        \subsubsection{Hybrid Approaches}
            Finally, we compare our model with the point process based approaches. These approaches are applicable to both evaluation setups. First, we test the check-in time prediction model presented in \cite{yang2018recurrent} (RSTPP). Here, an RNN is used to extract temporal features from the historical check-in times of users. Combining with the user trajectory information, the intensity is computed as
	
        	\begin{equation}
        		\lambda(t,\textbf{l}|\Omega(t)) = \text{exp}(\textbf{w}_\text{h}\textbf{h}_i + w_\text{t}(t-t_{i-1}) + w_\text{l}||\textbf{l} - \textbf{l}_{i-1}||^2 + b),
        	\end{equation}
        	where $ \textbf{h}_i $ is the state vector of the RNN model, which has the state transition
        	
        	\begin{equation}
        		\textbf{h}_i = \text{tanh}(\textbf{W}_\text{x}^T\textbf{x}_{i-1} + \textbf{W}_\text{h}^T\textbf{h}_{i-1}).
        	\end{equation}
        	Originally, user activity features and past check-in time and locations were included in the vector $ \textbf{x}_i $. However, we only use the past time and location information. 
        	
        	Similar to the check-in time prediction, Recurrent Marked Temporal Point Process (RMTPP) approach also estimates the intensity via an RNN \cite{du2016recurrent}. However, in this case, the intensity is defined only for the temporal axis and the location is predicted with markers. Both the intensity and the location markers are computed from the state vector of the RNN model.

    \subsection{Evaluation of the Presented Algorithm}
        In this section, we present the experiment results for assessing the presented algorithm on real-life datasets. We split the data sequence $ \{X_t\}_{t=1}^T $ into three sets: training, validation, and test sets, which are ordered with time. These sets have split boundary indices $ T_\text{train} $, $ T_\text{validation} $, and $ T $, respectively. Thus, the training, validation and test sets are defined as $ \textbf{X}_\text{train} = \{X_t\}_{t=1}^{T_\text{train}} $, $ \textbf{X}_\text{validation} = \{X_t\}_{t=T_\text{train}}^{T_\text{validation}} $, and $ \textbf{X}_\text{test} = \{X_t\}_{t=T_\text{validation}}^{T} $.

    \subsubsection{Prediction Horizon Effect}
\begin{figure*}[htp]%
        \centering
        \subfigure[\centering Prediction horizon RMSE comparison for the Chicago Crime Dataset]{{\includegraphics[width=0.47\textwidth]{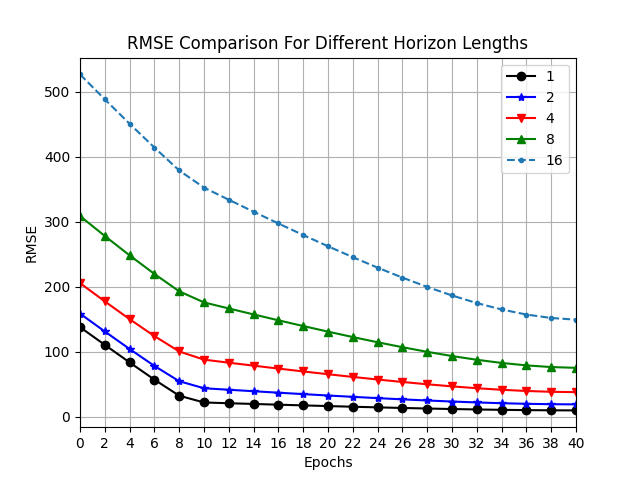}}
        \label{fig:horizon_chicago_mse}}%
        \subfigure[\centering Prediction horizon NLL comparison for the Chicago Crime Dataset]{{\includegraphics[width=0.47\textwidth]{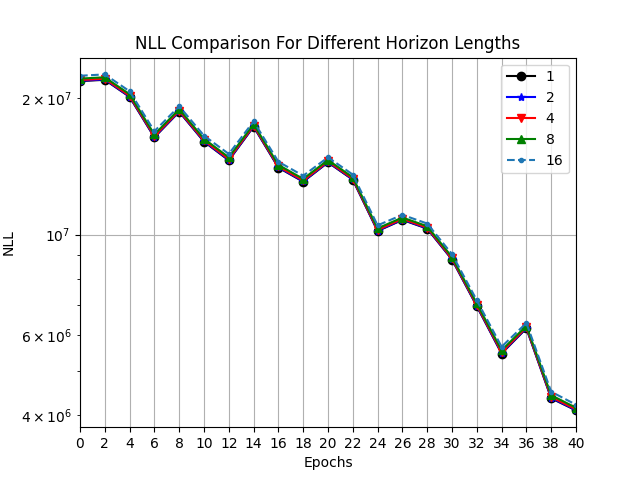}}
        \label{fig:horizon_chicago_nll}}\qquad
        \subfigure[\centering Prediction horizon RMSE comparison for the Simulated Dataset]{{\includegraphics[width=0.47\textwidth]{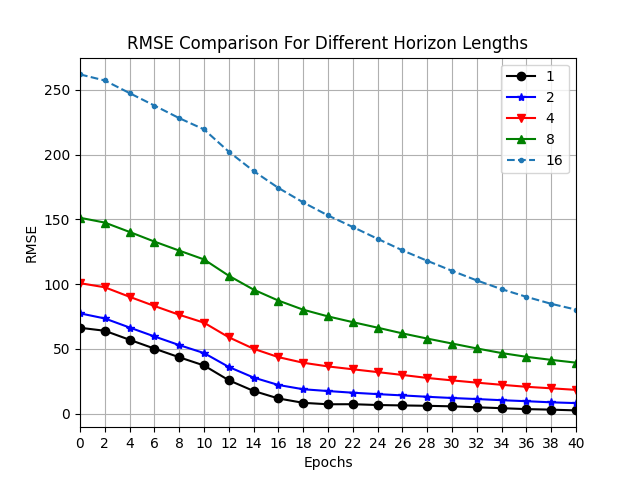}}
        \label{fig:horizon_simulated_mse}}%
        \subfigure[\centering Prediction horizon NLL comparison for the Simulated Dataset]{{\includegraphics[width=0.47\textwidth]{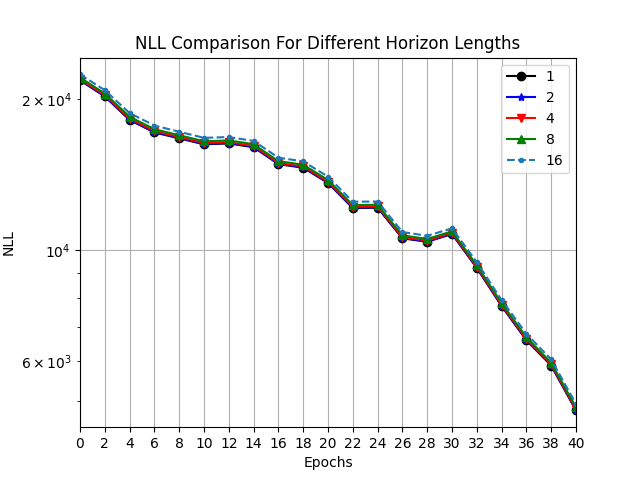}}
        \label{fig:horizon_simulated_nll}}%
        \caption{Prediction horizon performance comparison for our algorithm on two datasets.}%
        \label{fig:horizon_test}%
        \end{figure*}

        We investigate the performance for different prediction horizon lengths. For horizon length $h$, the prediction frame $ \hat{y}_t $ is generated by accumulating the intensity in the horizon interval. We have selected spatial and temporal resolutions as 3 km and 30 days. Furthermore, we choose the past window length as 30 days due to its performance. From Fig. \ref{fig:horizon_chicago_mse} and \ref{fig:horizon_simulated_mse}, we can observe that the error increases with the increasing horizon length. Moreover, we can see that the increase in error is not linearly increasing with the horizon length. A unit length of horizon in the figures correspond to 15 days.
        
        We assess the performance on three datasets: Chicago Crime dataset, Earthquake dataset and the simulated dataset. For the crime and simulated dataset, we compute the RMSE and the NLL in the validation sets through epochs as in Fig. \ref{fig:horizon_chicago_mse}, \ref{fig:horizon_chicago_nll}, \ref{fig:horizon_simulated_mse}, and \ref{fig:horizon_simulated_nll} respectively. 
        
        We also perform the same experiments on the earthquake dataset, which contains less events and has a sparse distribution. Finally, we report the test results for all datasets and all horizon lengths on Table \ref{table:horizon_test}. Furthermore, we also report the RMSE error when predicting number of events occurring in a unit spatio-temporal cell with the mean number of events per cell for comparison.
        
        \begin{table*}[]
        \begin{tabular}{|c|c|c|c|c|c|c|}
        \hline
        \textbf{\begin{tabular}[c]{@{}c@{}}Horizon (15 days)\end{tabular}} & \textbf{\begin{tabular}[c]{@{}c@{}}Chicago RMSE\end{tabular}} & \textbf{\begin{tabular}[c]{@{}c@{}}Chicago Average\end{tabular}} & \textbf{\begin{tabular}[c]{@{}c@{}}Simulated\\ RMSE\end{tabular}} & \textbf{\begin{tabular}[c]{@{}c@{}}Simulated\\ Average\end{tabular}}  & \textbf{\begin{tabular}[c]{@{}c@{}}Earthquake\\ RMSE\end{tabular}} & \textbf{\begin{tabular}[c]{@{}c@{}}Earthquake\\ Average\end{tabular}}  \\ \hline
        \textbf{1}                                                                   &                      10.21                                           &                               29.56                                 &                       3.12                                            &                                 5.37    &   4.12   & 13.12                                 \\ \hline
        \textbf{2}                                                                   &                      21.27                                         &                                 31.02                                 &                            7.61                                       &                                    10.74     &  7.01  & 26.24                         \\ \hline
        \textbf{4}                                                                   &                      39.15                                         &                             62.04                                   &                              17.74                                     &                               21.48          &   15.93  & 52.48                         \\ \hline
        \textbf{8}                                                                   &                      75.44                                             &                              124.08                                  &                         38.11                                          &                              42.96      &   38.20   & 104.96                              \\ \hline
        \textbf{16}                                                                  &                      148.32                                           &                              248.16                                  &                           78.12                                        &                              85.92       &   81.55  & 209.92                             \\ \hline
        \textbf{32}                                                                  &                      294.12                                           &                             496.32                                   &                           158.80                                        &                             171.84      &   157.49   & 419.84                               \\ \hline
        \textbf{64}                                                                  &                      569.46                                           &                             992.64                                   &                           313.78                                        &                             343.68      &   311.25   & 839.68                               \\ \hline
        \end{tabular}
        \caption{Experiment results on crime and simulated datasets for different horizon lengths.}
        \label{table:horizon_test}
        \end{table*}
        
        At the end of the training, the algorithm achieved an RMSE of $ 10.21 $ on the crime dataset, which is significantly less than $ 526.21 $, the average number of events observed in 15 day frame. We observe that the model achieves small improvements in the prediction performance after the horizon length 16 (240 days). After this point, the RMSE performance is close to the average event numbers. Moreover, we also observe that the performances are better for the experiments on simulated data, as we have generated the simulated data with the same model structure.
        
    \subsubsection{Past Temporal Window Length Effect}
\begin{figure*}[htp]%
        \centering
        \subfigure[\centering Past temporal window length RMSE comparison for the Chicago Crime Dataset]{{\includegraphics[width=0.32\textwidth]{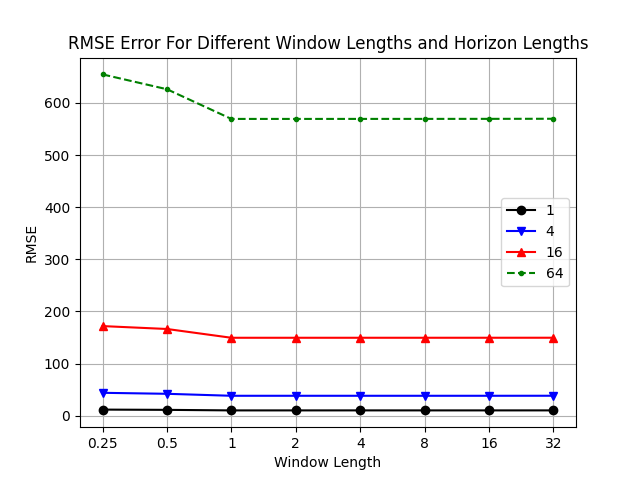}}
        \label{fig:window_chicago}}
        \subfigure[\centering Past temporal window length RMSE comparison for the Simulated Dataset]{{\includegraphics[width=0.32\textwidth]{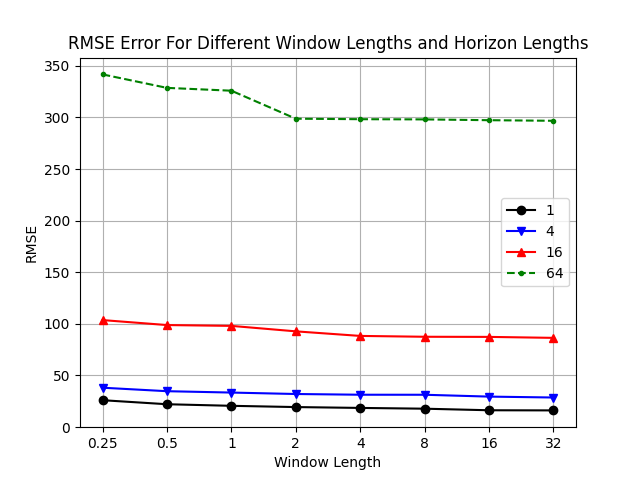}}
        \label{fig:window_simulated}}
        \subfigure[\centering Past temporal window length RMSE comparison for the Earthquake Dataset.]{{\includegraphics[width=0.32\textwidth]{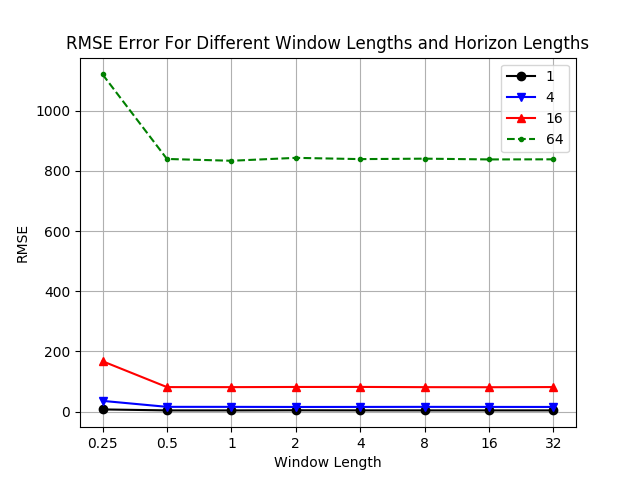}}
        \label{fig:window_earthquake}}
        \caption{Past temporal window length comparison for different horizon lengths on two datasets.}%
        \label{fig:window_test}%
        \end{figure*}

        We also measure the effect of choosing different temporal window lengths on the RMSE performance. We perform 4 experiments with window lengths 1, 4, 16, and 64. A unit of temporal length corresponds to 15 days as in the previous setup. For 4 window lengths, we measure the RMSE with 4 different horizon lengths on the crime, simulated and earthquake data as in Fig. \ref{fig:window_chicago}, \ref{fig:window_simulated}, and \ref{fig:window_earthquake}, respectively. 
        
        In all of the experiments, we can see that the RMSE converges to a level after a certain window length, which shows the temporal dependency length for each dataset. Specifically, for the crime dataset, we can see that choosing the window length 1 (15 days) is sufficient.
        
        We obtain the results in Fig. \ref{fig:window_earthquake} on the earthquake dataset. In this case, we can observe that the improvement is more apparent as the data consists of temporally spaced events. This is due to the collection of earthquakes that have magnitude higher than a certain level. This filters out preshocks and aftershocks, which eventually results in a sparse event sequence. Thus, to capture more past events, the time window should be selected longer, as it is also shown empirically in the figures.
        
        We also report the RMSE performances on Table \ref{table:crime_rmse_results} and \ref{table:earthquake_rmse_results}

    \subsubsection{Parallel and Separate Training}
\begin{figure*}[htp]%
        \centering
        \subfigure[\centering Parallel and Separate training RMSE comparison for the Chicago Crime Dataset]{{\includegraphics[width=0.32\textwidth]{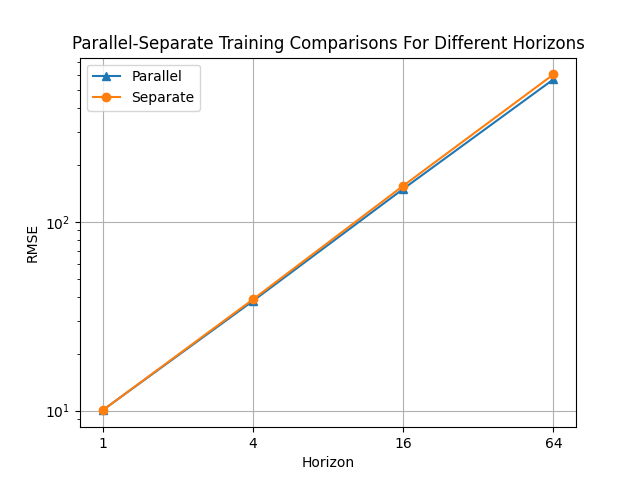}}
        \label{fig:parallel_chicago}}
        \subfigure[\centering Parallel and Separate training RMSE comparison for the Simulated Dataset]{{\includegraphics[width=0.32\textwidth]{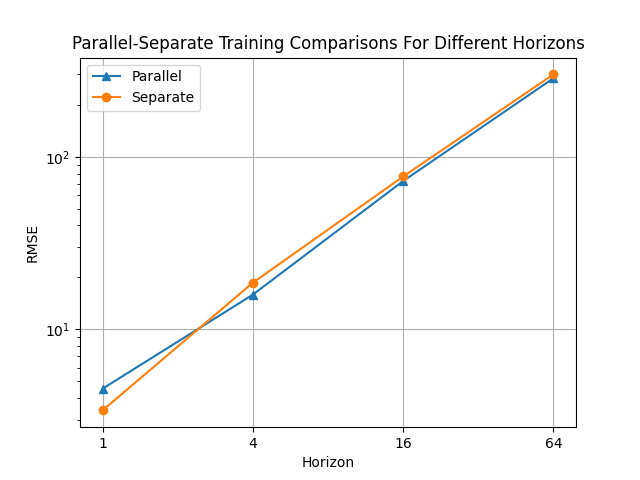}}
        \label{fig:parallel_simulated}}
        \subfigure[\centering Parallel and Separate training RMSE comparison for the Earthquake Dataset]{{\includegraphics[width=0.32\textwidth]{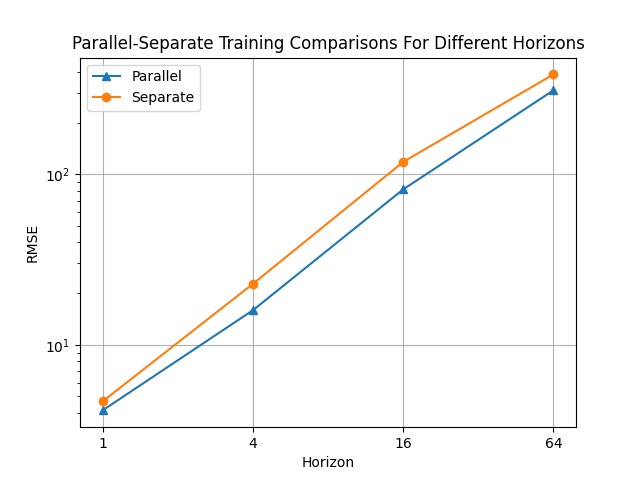}}
        \label{fig:parallel_earthquake}}
        \caption{Parallel and Separate training comparison on two different datasets.}%
        \label{fig:parallel_test}%
        \end{figure*}

        We perform two different training modes for different horizon lengths. In separate training mode, the model is trained with a single negative-log-likelihood loss function that is computed for a specific horizon length. In the parallel mode, the model is trained with the negative log-likelihood loss for all horizons at the same time. This method has positive impact on the model performance due to better generalization capability. Training the model for predicting the events in near and far future results in a better set of parameters that are capable of modeling the actual sequence. We show this with the empirical results in Fig. \ref{fig:parallel_chicago} and \ref{fig:parallel_simulated}, and \ref{fig:parallel_earthquake} for the crime, earthquake, and simulated data, respectively.
        
        In all of the datasets, the parallel training outperforms the separate training. For this reason, in all experiment sets, we use the parallel training mode. In addition to these results, we also present the performance on the test set in Table , where P-TreeHawkes and S-TreeHawkes correspond to the parallel and separate scenarios respectively.
        
    \subsubsection{Spatial Subregion Number Effect}
\begin{figure*}[htp]%
        \centering
        \subfigure[\centering RMSE comparison for different tree depths on the Chicago Crime Dataset]{{\includegraphics[width=0.48\textwidth]{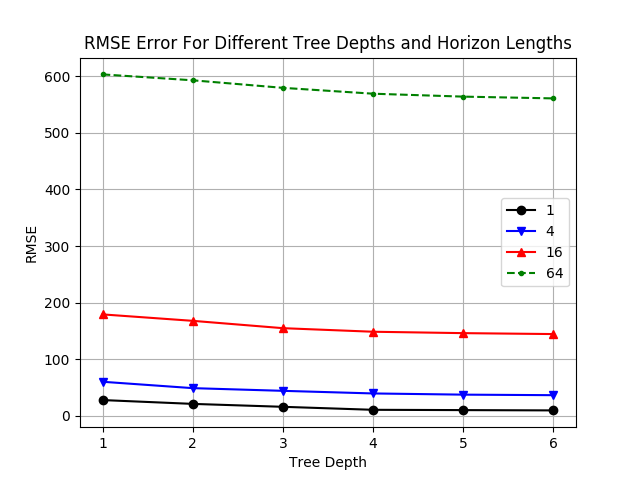}}
        \label{fig:chicago_region}}
        \subfigure[\centering RMSE comparison for different tree depths on the Earthquake Dataset]{{\includegraphics[width=0.48\textwidth]{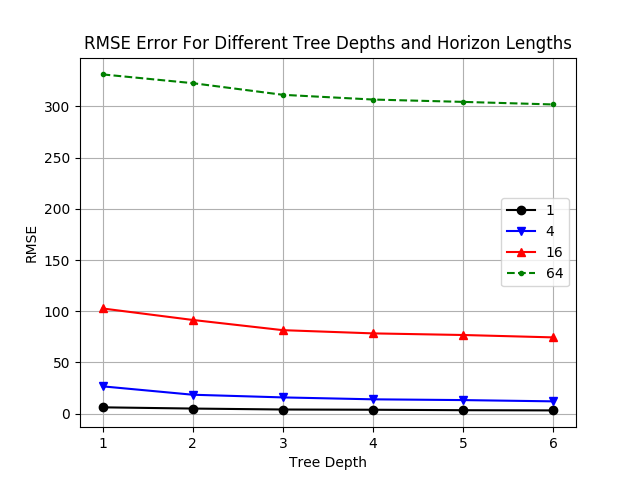}}
        \label{fig:earthquake_region}}
        \caption{Performance comparison for different number of spatial subregions on the crime and earthquake datasets.}
        \end{figure*}

        Since the presented algorithm consists of a decision tree that is used to partition the spatial region of interest, we experiment with different tree depths. As the depth of the tree increases, the number of spatial subregions increase exponentially. The spatial precision in the predictions will increase with the increasing number of subregions. We demonstrate the effect on the performance for different tree depths as in Fig. \ref{fig:chicago_region} and \ref{fig:earthquake_region}, respectively.
        
        From the results in the figures, as the number of spatial subregions increase, the model performance also increase. We observe that the performance converges to a level at the tree with depth 4 on the crime dataset. Thus, we use a 4-level tree in our experiments for comparison. We also perform the same set of experiments on the earthquake dataset. Although we observe a similar increase in the performance level with the increasing tree depth, 3 is selected for the tree depth on this dataset. Since the tree structure we are using is a binary tree, the effective number of regions for these scenarios are 16 and 8, respectively.

	\subsection{Performance Comparison With Benchmark Models}
	    RMSE comparisons for the crime dataset for different models are shown in Table \ref{table:crime_rmse_results}. In Table \ref{table:crime_rmse_results}, S-TreeHawkes corresponds to the separate training scenario and P-TreeHawkes corresponds to the parallel training scenario. It is clear that the performance of our algorithm surpasses the rest of the approaches on this dataset. However, the performance margin decreases with the increasing horizon length. This is due to the averaging effect for the event counts in long temporal intervals. As the temporal horizon length is increased, the number of events inside the horizon will converge to the average density of events in time. 
	    
	   For the crime dataset, we can see that the CNN model achieves the second best performance in higher horizon lengths after our algorithm. This is due to the capability of CNNs in capturing spatial patterns. Moreover, the CNN model is used in an auto-regressive manner, which allows it to learn temporal patterns in a fixed window length. The exceptional performance of our model and the CNN is due to the performance in modeling spatial interactions. Unlike the other methods that independently model spatial patterns, or incorporate the location information additively, the kernel mechanism in our model demonstrated an exceptional performance. The performance difference between our model and the CNN is due to the non-stationary modeling ability of our model in the spatial domain. Since the CNN applies the same kernel through the whole space, different spatial dynamics may not be effectively processed under different dynamics. On the other hand, we model each partition with an individual process.
	   
	   \begin{table}[]
	   \centering
        \begin{tabular}{|l|l|l|l|l|}
        \hline
        \textbf{Model}        & \textbf{1}    & \textbf{4}     & \textbf{16}     & \textbf{64}     \\ \hline
        \textbf{P-TreeHawkes} & 10.21         & \textbf{39.15} & \textbf{148.32} & \textbf{569.46} \\ \hline
        \textbf{S-TreeHawkes} & \textbf{9.95} & 40.83          & 154.41          & 582.21          \\ \hline
        \textbf{CNN}          & 11.52         & 46.64          & 167.36          & 604.75          \\ \hline
        \textbf{RMTPP}        & 15.43         & 57.11          & 182.98          & 622.34          \\ \hline
        \textbf{RSTP}         & 16.65         & 57.73          & 180.36          & 616.47          \\ \hline
        \end{tabular}
        \caption{RMSE Test results on the Chicago Crime Dataset}
        \label{table:crime_rmse_results}
        \end{table}
	   
	   RMSE comparisons for the earthquake dataset are shown in Table \ref{table:crime_rmse_results}. For this comparison, we can see that the convergence between performances is less apparent. This is due to the nature of the earthquake sequences, which are more sparsely distributed compared to criminal event records. 
	   
	   \begin{table}[]
	   \centering
        \begin{tabular}{|l|l|l|l|l|}
        \hline
        \textbf{Model}        & \textbf{1}    & \textbf{4}     & \textbf{16}    & \textbf{64}     \\ \hline
        \textbf{P-TreeHawkes} & \textbf{4.12} & \textbf{15.93} & \textbf{81.55} & \textbf{311.25} \\ \hline
        \textbf{S-TreeHawkes} & 4.55          & 23.40          & 119.26         & 411.63          \\ \hline
        \textbf{CNN}          & 15.36         & 57.31          & 167.47         & 468.11          \\ \hline
        \textbf{RMTPP}        & 7.38          & 23.62          & 126.89         & 430.27          \\ \hline
        \textbf{RSTP}         & 5.97          & 22.45          & 130.71         & 454.84          \\ \hline
        \end{tabular}
        \caption{RMSE Test results on the Earthquake Dataset}
        \label{table:earthquake_rmse_results}
        \end{table}
	   
	   We also observe that the point process based approaches outperform the CNN model. This results shows that the temporal patterns in certain regions have more significant effect compared to spatial patterns on the future event arrivals. However, both of the probabilistic approaches yield worse results compared to our model. Unlike the independent temporal and spatial location modeling, our model is capable of jointly modeling both temporal and spatial patterns.
	   
	   \begin{figure*}[htp]%
        \centering
        \subfigure[\centering RMSE comparison for different models on the time estimation on Earthquake Dataset]{{\includegraphics[width=0.49\textwidth]{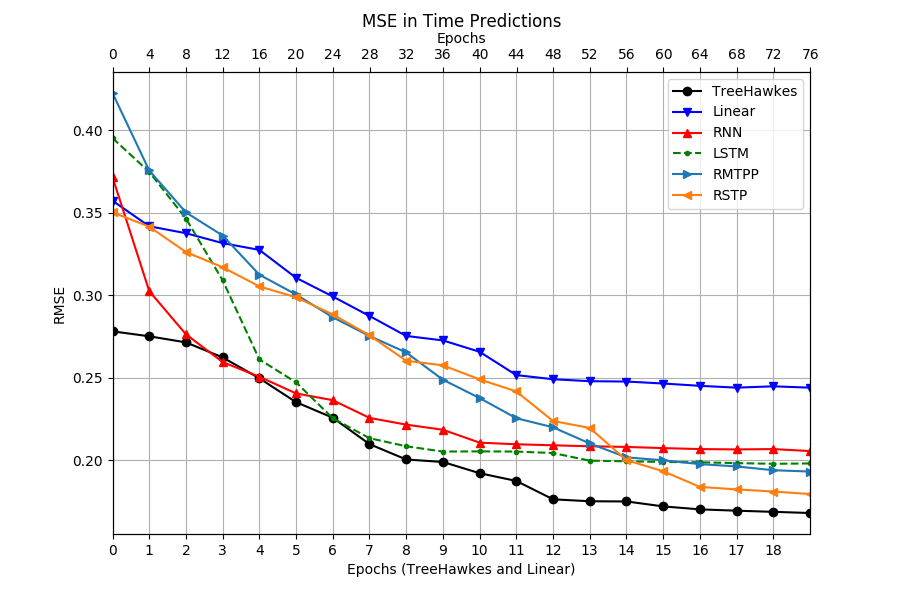}}
        \label{fig:earthquake_time}}
        \subfigure[\centering RMSE comparison for different models on the location estimation on Earthquake Dataset]{{\includegraphics[width=0.49\textwidth]{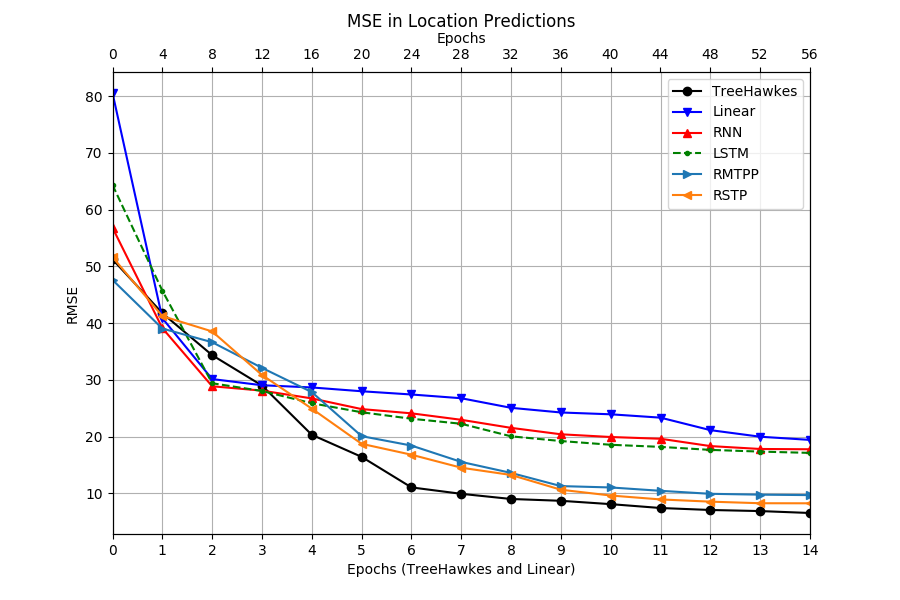}}
        \label{fig:earthquake_loc}}
        \caption{Performance comparison in both time and location estimations on the earthquake dataset.}
        \end{figure*}
	   
	   Finally, to complete our performance analysis, we also provide the results achieved for the time and location estimations of earthquake events in Fig. \ref{fig:earthquake_time} and \ref{fig:earthquake_loc} for the validation set. We also give the test results in Table \ref{table:final} for both time (MSE-t) and location (MSE-l) estimations. In both time and location estimations, our model achieves the best performance, which is followed by the point process based approaches. This is due to the capabilities of these methods for modeling sparse sequences. However, all methods weakly combine temporal and spatial information as they either independently model each domain, or incorporate information with a simple model such as a linear combination.
	   
	   \begin{table}[]
	   \centering
        \begin{tabular}{|l|l|l|}
        \hline
        \textbf{Models}     & \textbf{MSE-t} & \textbf{MSE-l} \\ \hline
        \textbf{TreeHawkes} & 0.162          & 6.47           \\ \hline
        \textbf{Linear}     & 0.271          & 33.28          \\ \hline
        \textbf{RNN}        & 0.202          & 31.86          \\ \hline
        \textbf{LSTM}       & 0.186          & 29.78          \\ \hline
        \textbf{CNN}        & 0.264          & 25.47          \\ \hline
        \textbf{RSTPP}      & 0.192          & 15.6           \\ \hline
        \textbf{RMTPP}      & 0.189          & 19.2           \\ \hline
        \end{tabular}
        \caption{Test results for the time and location estimations on the earthquake dataset.}
        \label{table:final}
        \end{table}
	   
	   As it can be seen from the figure, performance of the linear model is limited to a certain level both in time and location predictions. This is due to the simple structure of the model, which is a linear combination of the past sample times and locations. For the RNN and LSTM models, we obtain better results compared to the linear model both in time and location predictions. The reason behind this performance increase is due to the nonlinearity and the inherent state introduced by the RNN and LSTM transition equations. Instead of using a fixed number of past sample times and locations for prediction, this model computes the predictions using the state vector.

\section{Concluding Remarks}\label{sec:conclusion}

	We introduce a novel spatio-temporal prediction model that predicts the times and locations of the samples as well as the expected number of events in a spatio-temporal interval using the past samples. Our formulations are based on the Hawkes process, but can be readily extended to other point process models depending on the application. We also incorporate the spatial and temporal connections between the past samples and the intensity function via kernel mechanisms. Furthermore, we also partition the spatial region into subregions via an adaptive decision tree. Therefore, we optimize our point process parameters jointly with the subregion boundaries using a likelihood based objective function and the stochastic gradient descent method. Thanks to self-exciting and non-stationary intensity formulation of our point process model and the adaptive partitioning mechanism, we are able to represent highly sparse and non-stationary data. Finally, we demonstrate significant performance improvements through an extensive set of experiments where we compare our model with the baseline and standard approaches on a real-life crime and an earthquake dataset.

\begin{spacing}{.87}
\bibliographystyle{IEEEtran}
\bibliography{main}
\end{spacing}
\end{document}